\documentclass{article} % For LaTeX2e

\usepackage{iclr2027_conference,times}
\iclrfinalcopy
\usepackage{amsmath,amssymb}
\usepackage{amsmath,amsfonts,bm}

\def\eqref#1{equation~\ref{#1}}
\def\1{\bm{1}}

\DeclareMathAlphabet{\mathsfit}{\encodingdefault}{\sfdefault}{m}{sl}
\SetMathAlphabet{\mathsfit}{bold}{\encodingdefault}{\sfdefault}{bx}{n}

\usepackage[utf8]{inputenc}
\usepackage[T1]{fontenc}

\usepackage{graphicx}
\usepackage{svg}
\usepackage{wrapfig}
\usepackage{booktabs}
\usepackage[table]{xcolor}

\usepackage[hidelinks]{hyperref}
\usepackage{url}
\usepackage{cleveref}

\usepackage{tikz}
\usepackage{pgfplots}
\pgfplotsset{compat=1.18}
\usepgfplotslibrary{polar}

\usepackage{listings}
\lstdefinestyle{taxonomyprompt}{
  basicstyle=\ttfamily\fontsize{7}{8.5}\selectfont,
  columns=fullflexible,
  keepspaces=true,
  breaklines=true,
  breakatwhitespace=false,
  showstringspaces=false,
  mathescape=false,
  numbers=none,
  frame=none,
  linewidth=\linewidth,
  aboveskip=4pt,
  belowskip=8pt
}

\usepackage{comment}
\newcommand{\ourmodel}{ProtoLIP}
\title{ProtoLIP: From Sentence-Level to Object-Level Evidence Disentanglement}

\author{
Yan Zhu \\
Department of Computer Science \\
Tulane University \\
New Orleans, LA, USA \\
\texttt{yzhu27@tulane.edu}
\And
Yongbo Chen \\
Department of Computer Science \\
Tulane University \\
New Orleans, LA, USA
\AND
Zhengming Ding \\
Department of Computer Science \\
Tulane University \\
New Orleans, LA, USA
\And
Rebecca Faust \\
Department of Computer Science \\
Tulane University \\
New Orleans, LA, USA
}

\begin{document}

\maketitle

\begin{abstract}
Query-conditioned vision--language models enable fine-grained interpretation
by revealing which visual content supports a given textual query and how this
evidence changes across queries. However, semantically, sentence-level evidence
does not necessarily decompose into object-specific contributions, while
spatially, object-level evidence can remain entangled with co-occurring objects
and surrounding scene context. Across multiple VLM architectures and independent
benchmarks, we observe persistent object-level evidence entanglement. Moreover,
exposed evidence maps do not necessarily correspond to the evidence that
directly constitutes the model's prediction. To disentangle visual evidence at
both semantic and spatial levels, we introduce \textbf{ProtoLIP}, a lightweight
prototype-mediated evidence layer that organizes reusable visual prototypes
into text-derived semantic families and uses \emph{coarse-to-fine evidence
routing}, where semantic families constrain prototype eligibility and the
complete query determines fine-grained prototype contributions. Our studies
show that \ourmodel{} improves evidence localization and separation across query
granularities, achieving average relative gains of 29\% in Pointing and 43\% in
Energy across four object- and phrase-level OOD benchmarks. Its localization
gains also transfer to independently pretrained VLMs, with larger improvements
observed in several transfer settings. On the primary backbone,
\ourmodel{} also improves image--text matching discrimination while remaining
competitive with a spatially supervised grounding model in object-level
localization. Crucially, ProtoLIP constructs its image--text matching score
directly from localized prototype evidence, enabling exact decomposition across
prototypes, semantic families, and spatial evidence without spatial annotations
or backbone retraining.
\end{abstract}

\section{Introduction}
\label{sec:intro}

\begin{figure}[h]
    \vspace{-12pt}
    \centering
    \includegraphics[width=\linewidth]{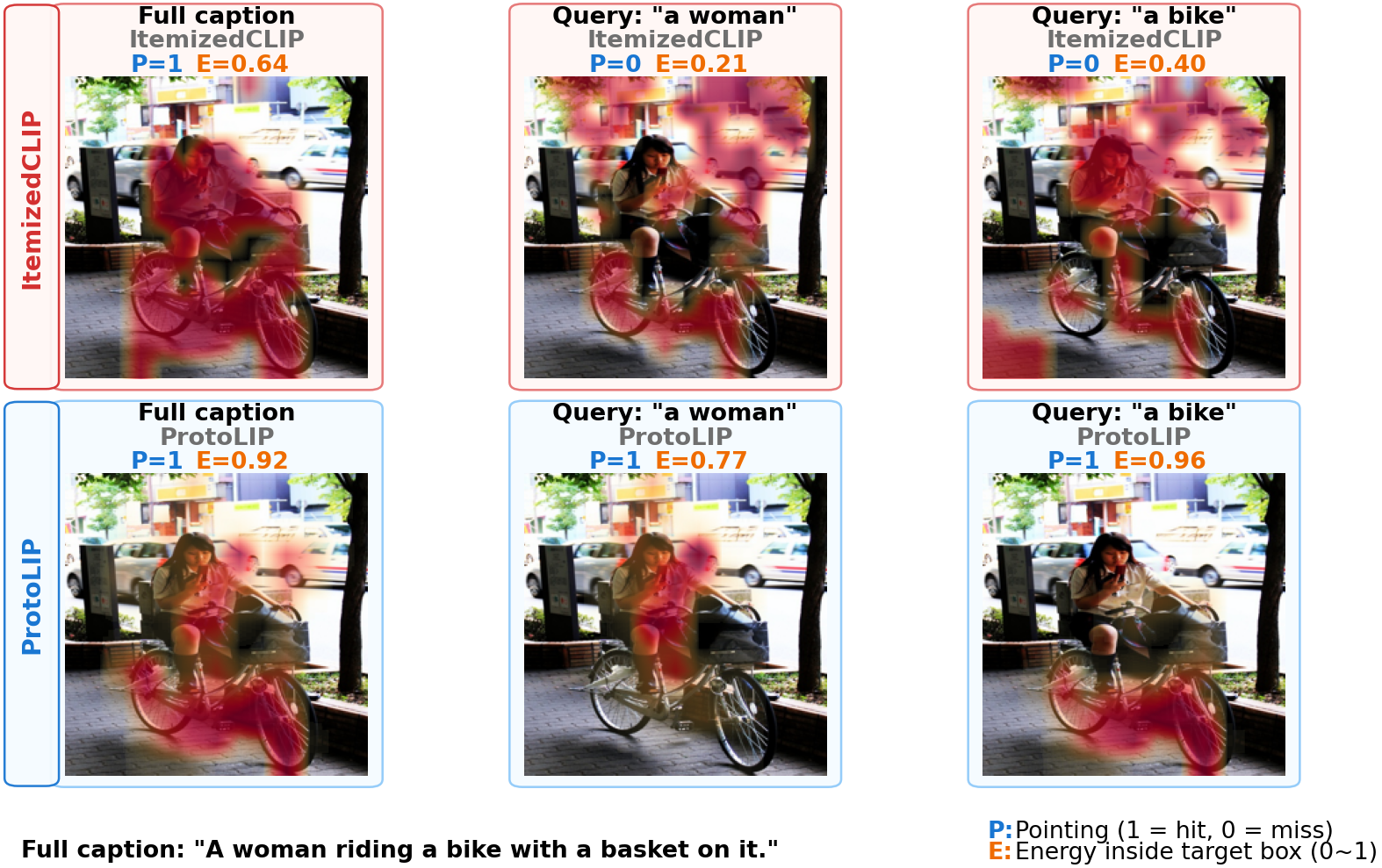}
    \caption{
    Object-level evidence disentanglement. ItemizedCLIP retains evidence
    from the woman and background when queried about objects,
    whereas ProtoLIP separates object-specific evidence.
    \textbf{P} indicates whether the peak evidence falls inside the target
    box, and \textbf{E} measures the fraction of evidence within it.
    }
    \label{fig:woman}
    \vspace{-12pt}
\end{figure}

Vision--language models (VLMs) learn transferable visual representations through image--text alignment~\citep{clip,siglip}. Recent models increasingly support fine-grained and text-conditioned visual representations, from long-caption and sub-caption supervision in DreamLIP~\citep{dreamlip}, to text-conditioned visual aggregation in FLAIR~\citep{flair}, and further to item-level independence and representation completeness in ItemizedCLIP~\citep{itemizedclip}. These advances enable the same image to be interpreted at different semantic granularities. Ideally, increasingly specific queries should yield correspondingly precise visual evidence, from complete descriptions to individual objects.
However, the objectives of existing fine-grained VLMs do not explicitly
require precise object-level visual evidence. ItemizedCLIP, for example, encourages distinct visual evidence for
sentence-level items containing multiple objects and relations, but does
not require object-level evidence within each sentence to be
disentangled. As
illustrated in Figure~\ref{fig:woman}, evidence for the woman and bicycle can
therefore remain entangled with each other and the surrounding context, even
under object-specific queries.

Recent studies report complementary evidence of this limitation.
SWIM~\citep{sun2026swim} shows that object nouns can produce diffuse visual
activations, while analyses of CLIP models identify reliance on background
regions and spurious contextual correlations
~\citep{agarwal2026conceptregions}. Related studies further identify
limitations in compositional concept binding and object-level localization
in pretrained VLMs
~\citep{lewis2024clip,bousselham2024grounding,assouel2025object}.
Together, these findings suggest that fine-grained textual conditioning
alone does not ensure object-specific visual evidence. We study this limitation as an evidence-disentanglement problem, asking
whether sentence-level evidence can be decomposed into distinct
object-specific contributions and spatially separated from co-occurring
entities and context.

We use \emph{evidence disentanglement} to denote the structured separation
of predictive evidence across semantic and spatial dimensions. We examine
two complementary aspects: separating evidence for a sentence into
contributions associated with its constituent objects, and spatially
separating object-specific evidence from co-occurring entities and context.
For this separation to explain a prediction, the evidence must also be
explicitly connected to the matching score. Localization alone does not
establish this connection: attention weights need not reflect predictive
influence~\citep{wu2024faithfulness,tokentm}, and gradient-based attribution
methods also face faithfulness and completeness
limitations~\citep{libragrad}.

To address these requirements, we introduce ProtoLIP, a lightweight
prototype-mediated evidence layer on top of a frozen VLM. ProtoLIP uses semantic families to select relevant prototypes and the full query to determine their contributions, constructing the image–text matching score directly from localized prototype evidence. This enables object-, phrase-, and sentence-level evidence while making the
resulting prediction exactly decomposable into prototype and semantic-family
contributions: these contributions are the additive terms of the matching
score itself, rather than a separate post-hoc explanatory layer.

We evaluate \ourmodel{} across query granularities, datasets, and
vision--language backbones, examining both evidence localization and
image--text matching. Our results show improved object-specific localization
and evidence separation across these settings, with localization gains
transferring to several independently pretrained VLM backbones. These gains
are achieved with a lightweight trainable layer on a frozen backbone. \ourmodel{} remains competitive with a spatially
supervised grounding model in object-level localization while improving
image--text matching on the primary backbone.

Our contributions are as follows.
\begin{itemize}
\item \ourmodel{}, a lightweight prototype-mediated evidence layer that
disentangles visual evidence at semantic and spatial levels while constructing
image--text matching predictions directly from localized prototype evidence,
enabling exact decomposition of the resulting prediction.

\item A comprehensive evaluation across datasets and VLM backbones, showing
improved object-specific localization, evidence separation, and image--text
matching on the primary backbone, with localization gains transferring across
independently pretrained VLMs.
\end{itemize}

\section{Related Work}
\label{sec:related}

\textbf{Fine-grained vision--language representations.}
Vision–language models such as CLIP~\citep{clip} learn transferable representations through contrastive image–text alignment, while SigLIP~\citep{siglip} replaces the softmax contrastive objective with pairwise sigmoid supervision. Subsequent work has pursued finer correspondence
between visual and textual content. FILIP~\citep{filip} introduces token-wise
interactions between visual and textual representations, while
RegionCLIP~\citep{regionclip} extends language--image alignment to localized
image regions. Other approaches enrich the textual supervision itself:
DreamLIP~\citep{dreamlip} leverages long captions and sub-captions to
learn more detailed visual--language correspondences.

More recent work makes the visual representation explicitly dependent on the
textual description. FLAIR~\citep{flair} uses text-conditioned attention
pooling over local visual tokens to produce language-informed representations
for fine-grained retrieval and localization. Most closely related to our
setting, ItemizedCLIP~\citep{itemizedclip} introduces item-independence and
representation-completeness objectives to learn distinct visual
representations for multiple semantically independent textual items describing
the same image.

Together, these approaches progressively refine the granularity of
vision--language alignment and textual conditioning. However, the unit of
textual supervision is not necessarily an individual visual object. Long
captions, sub-captions, and textual items may each contain multiple objects,
attributes, and relations; separating such descriptions therefore does not
require their constituent objects to have correspondingly separated visual
evidence. ProtoLIP addresses this remaining granularity gap by resolving rich textual descriptions into localized evidence for their individual objects.

\textbf{Vision--language localization and grounding.} A parallel line of work seeks to recover object-level spatial evidence from
pretrained VLM. Training-free methods such as NACLIP~\citep{naclip} derive dense localization maps
directly from pretrained CLIP representations without task-specific grounding
training. Dedicated grounding models such as Grounding
DINO~\citep{groundingdino} and MM Grounding DINO~\citep{mmgroundingdino}, in
contrast, use explicit detection and grounding supervision to learn
language-conditioned object localization. Our setting differs primarily in supervision. Dedicated grounding models rely
on explicit detection or grounding annotations, whereas ProtoLIP uses no spatial annotations and keeps the
pretrained VLM backbone frozen.

\textbf{Prototype-based visual evidence and prediction decomposition.}
Prototype-based models explain predictions through similarities to learned
representative patterns~\citep{protopnet}. Vision--language extensions include
class-specific visual prototypes for weakly supervised segmentation
(VPL~\citep{vpl}) and language-guided part prototypes for interpretable image
classification (PRISM~\citep{prism}). These methods primarily operate on
predefined categories, whereas \ourmodel{} targets compositional queries
containing multiple constituent objects.

More broadly, visually plausible attribution maps do not necessarily reflect
the evidence that actually determines a model's prediction~\citep{adebayo2018sanity}.
This issue also arises in prototype-based models, where prototype similarities
may directly contribute to a prediction while their spatial explanations are
not necessarily faithful to the underlying prototype computation
~\citep{wolf2024faithful}. In contrast, \ourmodel{} constructs the matching score directly
from localized prototype contributions, so that the resulting prediction is
exactly decomposable across prototypes and spatial evidence.

\begin{figure}[t]
\vspace{-18pt}
    \centering
    \includegraphics[width=\linewidth]{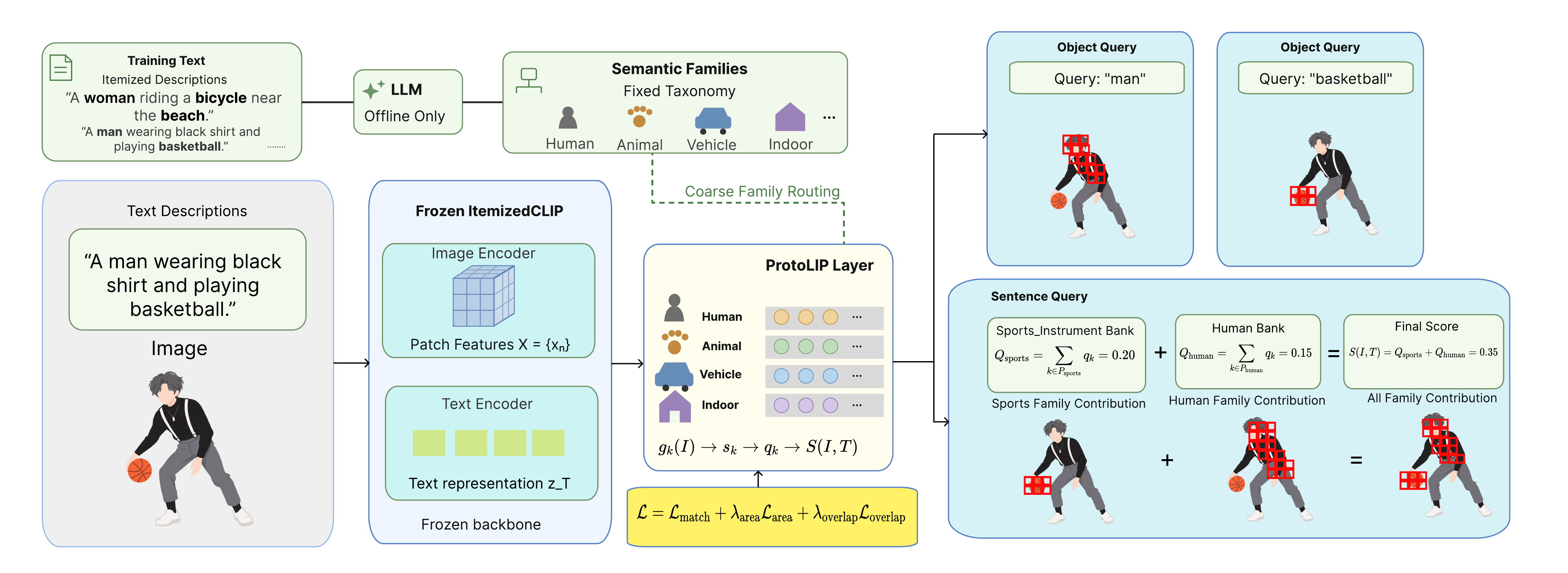}
    \vspace{-14pt}
    \caption{Overview of ProtoLIP. Training text is organized offline into coarse semantic families that constrain prototype eligibility. Given an image--text pair, a frozen vision--language backbone extracts patch features and a complete text representation. Eligible prototypes extract localized visual evidence, whose compatibility with the complete query determines their contribution to the matching score and query-specific spatial evidence.}
    \label{fig:overview}
    \vspace{-8pt}
\end{figure}
\section{Method}
\label{sec:method}

\subsection{Text-Derived Semantic Families}
\label{sec:family_construction}

Inspired by the class-specific prototype organization of
ProtoPNet~\citep{protopnet}, we introduce \textbf{coarse-to-fine evidence
routing}. Coarse semantic families first constrain prototype eligibility,
while the complete query then determines fine-grained prototype contributions.
This design provides semantic specialization without requiring the taxonomy
itself to encode fine-grained object distinctions. For example,
\emph{man}, \emph{woman}, and \emph{child} share the \emph{human} prototype
bank but can produce different contributions and spatial evidence.

We construct the taxonomy from training descriptions only. An offline LLM
extracts localizable visual concepts and aliases and groups them into coarse
families such as \emph{human}, \emph{vehicle}, and \emph{animal}, producing
a fixed dictionary
$\mathcal{D}=\{(F_c,\mathcal{A}_c)\}_{c=1}^{K}$, where $\mathcal{A}_c$
contains the concepts and aliases associated with family $F_c$. The LLM is
used only for this offline construction.

We use $M$ learnable visual prototypes
$\mathcal{P}=\{p_k\}_{k=1}^{M}$, where each prototype is a trainable vector
in the frozen backbone's patch-embedding space that learns to capture a
reusable visual pattern. The prototypes are learned during training by matching against image patches
using the objectives described in Section~\ref{sec:objective}.

We use $M=512$ prototypes distributed approximately uniformly across the
$K=39$ semantic-family banks, with 13--14 prototypes per family
(13.1 on average). Here, $K$ is determined by the offline
LLM-based taxonomy construction. During training and inference, we perform
\emph{semantic-family routing}: deterministic lexical matching activates
families whose aliases occur in the query, defining the eligible prototype
set $\mathcal{E}_T$. For example, ``a woman riding a bicycle'' activates the
\emph{human} and \emph{vehicle} families; if no family is matched, all
prototypes remain eligible.

Importantly, fine-grained evidence is determined by scoring eligible
prototypes with the complete query representation $z_T$. Thus, the taxonomy
acts only as a coarse semantic inductive bias, while object-specific evidence
remains query-dependent. We further analyze semantic-family routing coverage
on ADE20K, a challenging long-tail object benchmark, and Flickr30K Entities,
an open-vocabulary phrase-level benchmark. The largest localization gains
occur on matched queries, where semantic-family routing is active, while the
all-prototype fallback largely preserves localization performance on
unmatched queries. Full coverage statistics, matched/unmatched results, and
additional taxonomy and fallback analyses are provided in
Appendices~\ref{app:llm_prompts} and~\ref{app:natural_fallback}.

\subsection{Family-Constrained Prototype Evidence}
\label{sec:prototype}

Given the prototypes selected by semantic-family routing, we next determine
where each prototype extracts visual evidence and how strongly that evidence
supports the complete query. Each prototype first defines a spatial assignment
over the image patches, and the complete query then scores the resulting
prototype-specific evidence.
Let $p_k$ denote the $k$-th learned prototype,
$\operatorname{sim}(\cdot,\cdot)$ the feature-similarity function, and
$\tau_p$ the prototype temperature. Each prototype defines a normalized
spatial assignment
\begin{equation}
\beta_{k,n}(I)
=
\operatorname{softmax}_{n}
\left(
\frac{\operatorname{sim}(p_k,x_n)}{\tau_p}
\right),
\qquad
\sum_{n=1}^{N}\beta_{k,n}(I)=1.
\label{eq:prototype_assignment}
\end{equation}
The assignment $\beta_{k,n}$ determines where prototype $k$ extracts visual
evidence from the image. We use it to pool patch features into
prototype-specific visual evidence:
\begin{equation}
\widetilde{g}_k(I)
=
\sum_{n=1}^{N}\beta_{k,n}(I)x_n,
\qquad
g_k(I)
=
\phi_\theta\!\left(\widetilde{g}_k(I),p_k\right),
\label{eq:prototype_evidence}
\end{equation}
where $\widetilde{g}_k(I)$ is the pooled evidence and $\phi_\theta$ is a
lightweight gated residual MLP with learnable parameters $\theta$. Thus, $g_k(I)$ represents the image-specific evidence captured by prototype $k$.
Implementation details of the prototype evidence layer are provided in
Appendix~\ref{app:implementation}.

With the prototype evidence extracted, semantic-family routing determines which prototypes are eligible to
contribute. Each prototype has a learned positive weight that captures its relative
importance within the eligible prototype set
$r_k=\operatorname{softplus}(\rho_k)$. For an eligible prototype
$k\in\mathcal{E}_T$, we set
$\alpha_k(T)=r_k/\sum_{j\in\mathcal{E}_T}r_j$, and set
$\alpha_k(T)=0$ otherwise. The complete query embedding $z_T$ then evaluates
the compatibility of each eligible prototype's visual evidence:
\begin{equation}
s_k(I,T)=\operatorname{sim}(g_k(I),z_T),
\qquad
q_k(I,T)=\alpha_k(T)s_k(I,T),
\qquad
S(I,T)=\sum_k q_k(I,T).
\label{eq:prototype_score}
\end{equation}
Because the resulting contributions $q_k$ sum directly to $S(I,T)$, the
prototype-mediated matching prediction is exactly decomposable into
prototype contributions.

\subsection{Query-Specific Prediction Evidence}
\label{sec:query_evidence}

The previous step gives each prototype both a contribution to the matching
score and a spatial assignment over the image. We combine these quantities
to expose where the prediction is spatially supported while preserving the
complete prototype-mediated score:
\begin{equation}
C(n;I,T)
=
\sum_k q_k(I,T)\beta_{k,n}(I),
\qquad
\sum_{n=1}^{N}C(n;I,T)=S(I,T).
\label{eq:signed_query_evidence}
\end{equation}
The equality follows from the normalization of $\beta_k$. Thus, $C$ provides
a score-preserving and exhaustive spatial decomposition of the
prototype-mediated prediction: every contribution to $S(I,T)$ is assigned
to spatial evidence in $C$, with no residual score outside the exposed
evidence. Positive and negative values respectively support and oppose the
image--text match.

\begin{figure}[t]
 \vspace{-13pt}
    \centering
    \includegraphics[width=0.9\linewidth]{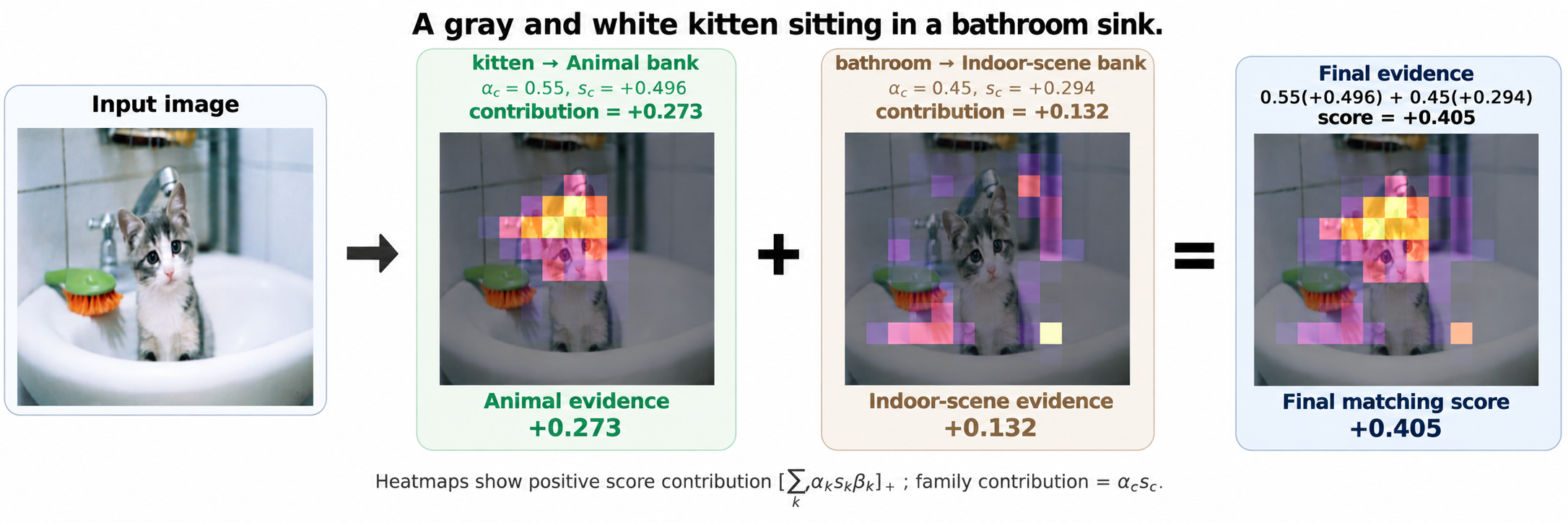}
    \vspace{-4mm}\caption{Exact decomposition of prototype evidence for a compositional sentence query.}
    \label{fig:evidence_composition}
    \vspace{-14pt}
\end{figure}

For localization and visualization, we normalize the positive part of $C$
to unit mass, denoted $H$, producing a nonnegative spatial evidence map.
Figure~\ref{fig:evidence_composition} illustrates this decomposition for a
compositional sentence query, showing distinct contributions from the
\emph{animal} and \emph{indoor-scene} families.

\subsection{Spatial Evidence Regularization}
\label{sec:spatial_evidence}

The prototype assignments above localize evidence, but do not encourage evidence from different semantic families to be compact and
spatially separated. We therefore regularize the spatial assignments of
active families to discourage diffuse evidence and excessive sharing between
families.

For a query with at least one matched family, let $\mathcal{F}_T$ denote the
active semantic families and $\mathcal{P}_c$ the prototypes assigned to
family $F_c$. We aggregate the prototype assignments within each active
family into the normalized family map
\begin{equation}
M_c(n;I,T)
=
\frac{\sum_{k\in\mathcal{P}_c}
\alpha_k(T)\beta_{k,n}(I)}
{\sum_{k\in\mathcal{P}_c}\alpha_k(T)}.
\label{eq:family_map}
\end{equation}
This map captures where the evidence associated with family $F_c$ is
spatially concentrated. We measure its spatial spread and the overlap between
two active families as
\begin{align}
A(M_c)
&=
\frac{1}{N}
\exp\!\left(
-\sum_{n=1}^{N}M_c(n)\log(M_c(n)+\epsilon)
\right),\\
O(M_c,M_{c'})
&=
\sum_{n=1}^{N}
\min\!\left(M_c(n),M_{c'}(n)\right),
\label{eq:spatial_measures}
\end{align}
where $\epsilon>0$ ensures numerical stability.
$A(M_c)$ is an entropy-based effective-area measure, while
$O(M_c,M_{c'})$ measures shared spatial mass.

Let $\mathcal{R}_T$ denote the unordered pairs of distinct active families.
We penalize only excessive area:
\begin{equation}
\mathcal{L}_{\mathrm{area}}
=
\frac{1}{|\mathcal{F}_T|}
\sum_{c\in\mathcal{F}_T}
[A(M_c)-\tau_a]_+^2.
\label{eq:area_loss}
\end{equation}
When $\mathcal{R}_T$ is nonempty, we additionally penalize excessive
pairwise overlap:
\begin{equation}
\mathcal{L}_{\mathrm{overlap}}
=
\frac{1}{|\mathcal{R}_T|}
\sum_{(c,c')\in\mathcal{R}_T}
[O(M_c,M_{c'})-\tau_o]_+^2.
\label{eq:overlap_loss}
\end{equation}
Here, $[u]_+=\max(0,u)$, and $\tau_a$ and $\tau_o$ are the tolerated area
and overlap thresholds. These losses encourage compact and separated family
evidence without forcing semantically related families to occupy disjoint
regions.

For single-family queries, $\mathcal{L}_{\mathrm{overlap}}=0$; unmatched
queries use all prototypes and are excluded from spatial regularization.
Within-family distinctions instead rely on query-conditioned prototype scoring.

\subsection{Learning Objective}
\label{sec:objective}

We jointly optimize image--text matching and the spatial organization of the
prototype evidence defined above. The objective combines a symmetric
contrastive matching loss with the two spatial regularizers.

For a batch of $B$ corresponding image--text pairs, let
$\mathbf{S}\in\mathbb{R}^{B\times B}$ with
$\mathbf{S}_{ij}=S(I_i,T_j)$ denote the pairwise score matrix and
$\mathbf{y}=(1,\ldots,B)$ the diagonal matching targets. The matching loss is
\begin{equation}
\mathcal{L}_{\mathrm{match}}
=
\frac{1}{2}
\left[
\operatorname{CE}(\mathbf{S}/\tau_s,\mathbf{y})
+
\operatorname{CE}(\mathbf{S}^{\top}/\tau_s,\mathbf{y})
\right],
\label{eq:matching_loss}
\end{equation}
where $\operatorname{CE}(\cdot,\cdot)$ denotes cross-entropy and
$\tau_s>0$ is the score temperature.

The complete objective is
\begin{equation}
\mathcal{L}
=
\mathcal{L}_{\mathrm{match}}
+
\lambda_{\mathrm{area}}\mathcal{L}_{\mathrm{area}}
+
\lambda_{\mathrm{overlap}}\mathcal{L}_{\mathrm{overlap}},
\label{eq:overall_loss}
\end{equation}
where $\lambda_{\mathrm{area}}$ and $\lambda_{\mathrm{overlap}}$ weight the
two spatial regularizers.

Spatial regularization is applied only to positive diagonal pairs, while the
same score $S(I,T)$ is used for training and inference.

\begin{figure}[h]
 \vspace{-11pt}
    \centering
    \includegraphics[width=0.98\linewidth]{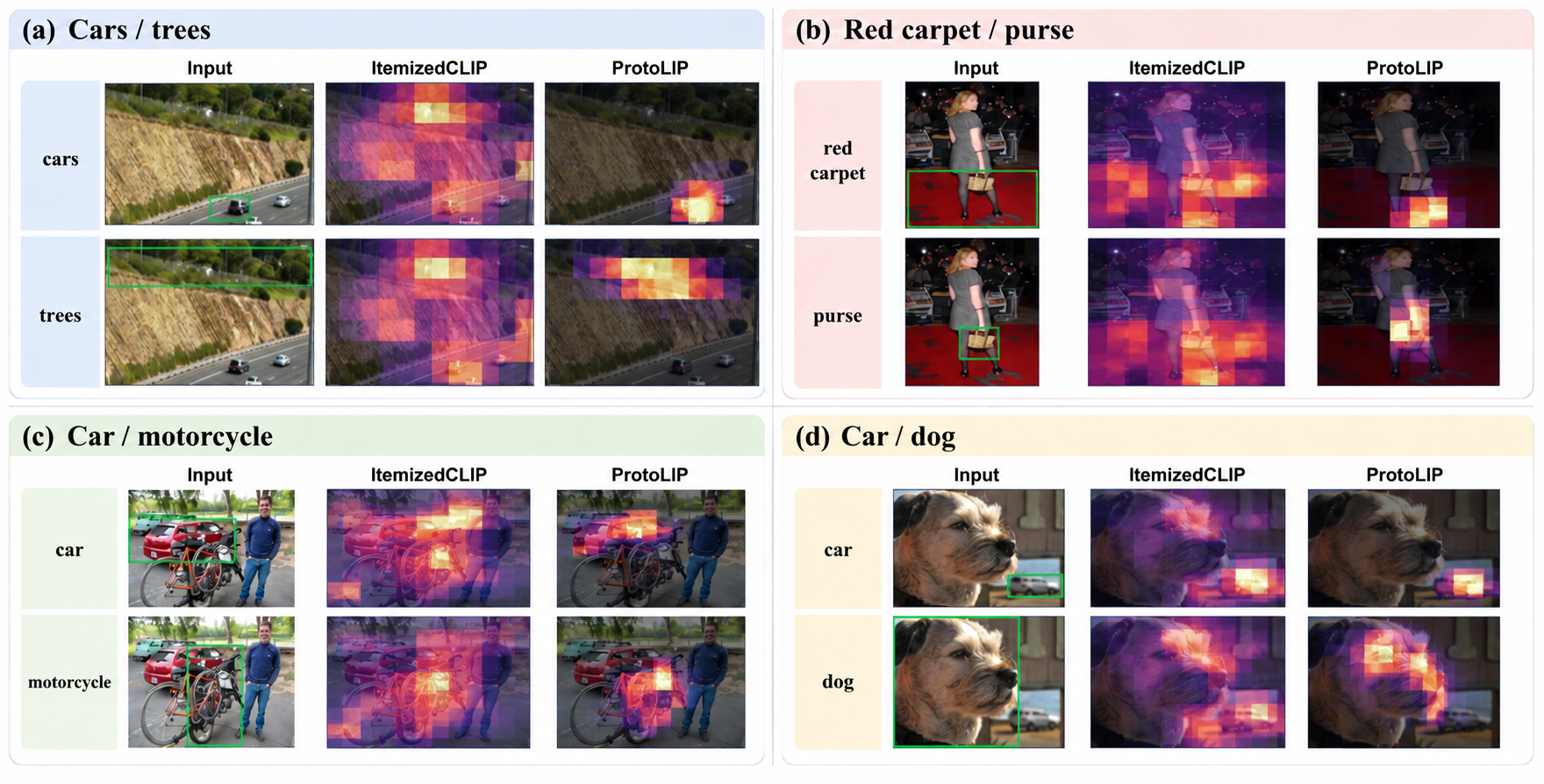}
    \vspace{-15pt}
\caption{
\textbf{Object-level evidence disentanglement.}
ItemizedCLIP entangles evidence across object queries, whereas
ProtoLIP localizes the queried object. Green boxes denote ground-truth regions.
}
    \label{fig:qualitative}
    \vspace{-11pt}
\end{figure}
\section{Experiments}
\label{sec:results}

\subsection{Experimental Setup}
\label{sec:setup}

\paragraph{Training and evaluation.}
We train \ourmodel{} on Itemized-CC0.3M~\citep{itemizedclip} with a frozen
ItemizedCLIP backbone, optimizing only the prototype evidence layer. All
external benchmarks are evaluated zero-shot. We evaluate object-level evidence
on COCO~\citep{coco}, VOC20~\citep{voc}, and ADE20K-150 validation
split~\citep{ADE20K}; phrase-level evidence on Flickr30K
Entities~\citep{flickr30k}; and caption-level evidence on COCO and Flickr30K
Entities. Because COCO lacks phrase-to-box links, we construct a caption-linked
object-query benchmark by matching explicit category mentions in captions to
COCO categories and treating all annotated instances of the matched category
as the target region. All methods use the same fixed manifest. Additional COCO-80 evaluation is provided in
Appendix~\ref{app:coco80}; full implementation, benchmark, and evaluation
details are provided in Appendix~\ref{app:implementation}.
\paragraph{Baselines and metrics.}
We compare against the frozen ItemizedCLIP backbone, released VLMs,
training-free localization methods, and spatially supervised grounding models
(Figure~\ref{fig:public_localization_curve}). We evaluate spatial evidence
using Pointing Game Accuracy~\citep{pointinggame}, which measures whether the
maximum-evidence location falls inside the ground-truth region, and the
Energy-Based Pointing Game (Energy)~\citep{scorecam}, which measures the
fraction of total positive evidence contained within the ground-truth region.
We evaluate image--text matching using area under the ROC curve (AUC) and
balanced accuracy (BAcc), and retrieval using Recall@k (I@k/T@k).

\subsection{Evidence Disentanglement Across Query Granularities}
\label{sec:evidence_disentanglement}

We first test whether \ourmodel{} can mitigate the object-level evidence
entanglement illustrated in Figures~\ref{fig:woman} and~\ref{fig:qualitative} on OOD benchmarks. We evaluate whether evidence
for an object-level query is better localized to the annotated region of the
queried object, while also measuring whether this improved localization is
accompanied by stronger image--text matching.

\begin{figure*}[h]

\centering

\definecolor{baseblue}{RGB}{120,145,162}
\definecolor{protolipred}{RGB}{225,95,95}

% ============================================================
% LEFT: Spatial Evidence Disentanglement
% ============================================================
\begin{minipage}[t]{0.64\textwidth}
\vspace{-5pt}
\centering

\textbf{(a) Spatial Evidence Disentanglement}

\vspace{3pt}

\begin{tikzpicture}
\begin{axis}[
ybar,
width=\linewidth,
height=5.2cm,
ymin=0,
ymax=1.0,
ylabel={Score},
ytick={0,0.2,0.4,0.6,0.8,1.0},
xtick={1,2,3,4,5,6,7,8,9,10,11,12},
xticklabels={P,E,P,E,P,E,P,E,P,E,P,E},
x tick label style={font=\scriptsize,yshift=-1pt},
yticklabel style={font=\scriptsize},
ylabel style={font=\scriptsize},
legend style={
    at={(0.5,1.03)},
    anchor=south,
    legend columns=2,
    font=\scriptsize,
    draw=none,
    /tikz/every even column/.append style={column sep=8pt}
},
enlarge x limits=0.025,
axis x line*=bottom,
axis y line*=left,
tick align=outside,
clip=false
]

% 
% ============================================================
% ProtoLIP: narrower foreground bars
% ============================================================
\addplot[
ybar,
bar width=7pt,
bar shift=0pt,
fill=protolipred,
draw=protolipred,
fill opacity=0.92,
nodes near coords={
    \pgfmathprintnumber[
        fixed,
        precision=2,
        zerofill
    ]{\pgfplotspointmeta}
},
nodes near coords style={
    font=\fontsize{5}{5.5}\selectfont\bfseries,
    text=protolipred,
    anchor=south,
    yshift=1pt
}
] coordinates {
(1,0.765)
(2,0.709)
(3,0.394)
(4,0.353)
(5,0.594)
(6,0.515)
(7,0.353)
(8,0.325)
(9,0.454)
(10,0.427)
(11,0.841)
(12,0.804)
};

\legend{ProtoLIP, ItemizedCLIP}

\node[
    anchor=west,
    font=\tiny
] at (axis description cs:0.74,1.03)
{P: Pointing, E: Energy};

% ============================================================
% ProtoLIP: mean +/- std over three seeds
% ============================================================

% VOC20
\draw[black,line width=0.4pt]
(axis cs:1,0.763) -- (axis cs:1,0.767);
\draw[black,line width=0.4pt]
(axis cs:0.94,0.763) -- (axis cs:1.06,0.763);
\draw[black,line width=0.4pt]
(axis cs:0.94,0.767) -- (axis cs:1.06,0.767);

\draw[black,line width=0.4pt]
(axis cs:2,0.705) -- (axis cs:2,0.713);
\draw[black,line width=0.4pt]
(axis cs:1.94,0.705) -- (axis cs:2.06,0.705);
\draw[black,line width=0.4pt]
(axis cs:1.94,0.713) -- (axis cs:2.06,0.713);

\draw[black,line width=0.4pt]
(axis cs:2,0.709) -- (axis cs:2,0.709);
\draw[black,line width=0.4pt]
(axis cs:1.94,0.709) -- (axis cs:2.06,0.709);
\draw[black,line width=0.4pt]
(axis cs:1.94,0.709) -- (axis cs:2.06,0.709);

% ADE20K
\draw[black,line width=0.4pt]
(axis cs:3,0.388) -- (axis cs:3,0.400);
\draw[black,line width=0.4pt]
(axis cs:2.94,0.388) -- (axis cs:3.06,0.388);
\draw[black,line width=0.4pt]
(axis cs:2.94,0.400) -- (axis cs:3.06,0.400);

\draw[black,line width=0.4pt]
(axis cs:4,0.349) -- (axis cs:4,0.357);
\draw[black,line width=0.4pt]
(axis cs:3.94,0.349) -- (axis cs:4.06,0.349);
\draw[black,line width=0.4pt]
(axis cs:3.94,0.357) -- (axis cs:4.06,0.357);

% COCO Object / Phrase
\draw[black,line width=0.4pt]
(axis cs:5,0.5861) -- (axis cs:5,0.6019);
\draw[black,line width=0.4pt]
(axis cs:4.94,0.5861) -- (axis cs:5.06,0.5861);
\draw[black,line width=0.4pt]
(axis cs:4.94,0.6019) -- (axis cs:5.06,0.6019);

\draw[black,line width=0.4pt]
(axis cs:6,0.5127) -- (axis cs:6,0.5173);
\draw[black,line width=0.4pt]
(axis cs:5.94,0.5127) -- (axis cs:6.06,0.5127);
\draw[black,line width=0.4pt]
(axis cs:5.94,0.5173) -- (axis cs:6.06,0.5173);

% COCO Full Caption
\draw[black,line width=0.4pt]
(axis cs:7,0.345) -- (axis cs:7,0.361);
\draw[black,line width=0.4pt]
(axis cs:6.94,0.345) -- (axis cs:7.06,0.345);
\draw[black,line width=0.4pt]
(axis cs:6.94,0.361) -- (axis cs:7.06,0.361);

\draw[black,line width=0.4pt]
(axis cs:8,0.323) -- (axis cs:8,0.327);
\draw[black,line width=0.4pt]
(axis cs:7.94,0.323) -- (axis cs:8.06,0.323);
\draw[black,line width=0.4pt]
(axis cs:7.94,0.327) -- (axis cs:8.06,0.327);

% Flickr Object / Phrase
\draw[black,line width=0.4pt]
(axis cs:9,0.4497) -- (axis cs:9,0.4583);
\draw[black,line width=0.4pt]
(axis cs:8.94,0.4497) -- (axis cs:9.06,0.4497);
\draw[black,line width=0.4pt]
(axis cs:8.94,0.4583) -- (axis cs:9.06,0.4583);

\draw[black,line width=0.4pt]
(axis cs:10,0.4251) -- (axis cs:10,0.4289);
\draw[black,line width=0.4pt]
(axis cs:9.94,0.4251) -- (axis cs:10.06,0.4251);
\draw[black,line width=0.4pt]
(axis cs:9.94,0.4289) -- (axis cs:10.06,0.4289);

% Flickr Full Caption
\draw[black,line width=0.4pt]
(axis cs:11,0.8314) -- (axis cs:11,0.8506);
\draw[black,line width=0.4pt]
(axis cs:10.94,0.8314) -- (axis cs:11.06,0.8314);
\draw[black,line width=0.4pt]
(axis cs:10.94,0.8506) -- (axis cs:11.06,0.8506);

\draw[black,line width=0.4pt]
(axis cs:12,0.8021) -- (axis cs:12,0.8059);
\draw[black,line width=0.4pt]
(axis cs:11.94,0.8021) -- (axis cs:12.06,0.8021);
\draw[black,line width=0.4pt]
(axis cs:11.94,0.8059) -- (axis cs:12.06,0.8059);

% ItemizedCLIP: wider background bars
% ============================================================
\addplot[
ybar,
bar width=10pt,
bar shift=0pt,
fill=baseblue,
draw=baseblue,
fill opacity=0.72,
nodes near coords={
    \pgfmathprintnumber[
        fixed,
        precision=2,
        zerofill
    ]{\pgfplotspointmeta}
},
nodes near coords style={
    font=\fontsize{5}{5.5}\selectfont,
    text=black,
    anchor=north,
    yshift=-2pt
}
] coordinates {
(1,0.538)
(2,0.465)
(3,0.343)
(4,0.279)
(5,0.426)
(6,0.332)
(7,0.316)
(8,0.249)
(9,0.378)
(10,0.308)
(11,0.837)
(12,0.703)
};

% ============================================================
% Dataset boundaries
% VOC20 | ADE20K | COCO | Flickr30K
% ============================================================
\draw[densely dashed,gray!55]
(axis cs:2.5,-0.3) -- (axis cs:2.5,1.0);

\draw[densely dashed,gray!55]
(axis cs:4.5,-0.3) -- (axis cs:4.5,1.0);

\draw[densely dashed,gray!55]
(axis cs:8.5,-0.3) -- (axis cs:8.5,1.0);

% ============================================================
% Query granularity boxes
% ============================================================

% COCO - Object / Phrase
\node[
anchor=north,
font=\fontsize{5}{5}\selectfont,
fill=gray!6,
minimum width=1.25cm,
minimum height=11pt,
inner sep=1pt
] at ([yshift=-13pt]axis cs:5.55,0)
{Object / Phrase};

% COCO - Full Caption
\node[
anchor=north,
font=\tiny,
fill=gray!6,
minimum width=1.25cm,
minimum height=11pt,
inner sep=1pt
] at ([yshift=-13pt]axis cs:7.45,0)
{Full Caption};

% Flickr - Object / Phrase
\node[
anchor=north,
font=\fontsize{5}{5}\selectfont,
fill=gray!6,
minimum width=1.275cm,
minimum height=11pt,
inner sep=1pt
] at ([yshift=-13pt]axis cs:9.55,0)
{Object / Phrase};

% Flickr - Full Caption
\node[
anchor=north,
font=\tiny,
fill=gray!6,
minimum width=1.275cm,
minimum height=11pt,
inner sep=1pt
] at ([yshift=-13pt]axis cs:11.45,0)
{Full Caption};

% ============================================================
% Dashed separators between query granularities
% ============================================================

% COCO: Object / Phrase | Full Caption
\draw[densely dashed,gray!55]
([yshift=-24pt]axis cs:6.5,0)
--
(axis cs:6.5,1.0);

% Flickr: Object / Phrase | Full Caption
\draw[densely dashed,gray!55]
([yshift=-24pt]axis cs:10.5,0)
--
(axis cs:10.5,1.0);
% ============================================================
% Dataset boxes
% ============================================================

\node[
anchor=north,
align=center,
font=\scriptsize\bfseries,
fill=gray!10,
draw=gray!35,
minimum width=1.35cm,
minimum height=20pt,
inner sep=2pt
] at ([yshift=-24pt]axis cs:1.5,0)
{VOC20\\[-1pt]\normalfont\tiny Object};

\node[
anchor=north,
align=center,
font=\scriptsize\bfseries,
fill=gray!10,
draw=gray!35,
minimum width=1.3cm,
minimum height=20pt,
inner sep=2pt
] at ([yshift=-24pt]axis cs:3.5,0)
{ADE20K\\[-1pt]\normalfont\tiny Object};

\node[
anchor=north,
font=\scriptsize\bfseries,
fill=gray!10,
draw=gray!35,
minimum width=2.5cm,
minimum height=20pt,
inner sep=2pt
] at ([yshift=-24pt]axis cs:6.5,0)
{COCO};

\node[
anchor=north,
font=\scriptsize\bfseries,
fill=gray!10,
draw=gray!35,
minimum width=2.55cm,
minimum height=20pt,
inner sep=2pt
] at ([yshift=-24pt]axis cs:10.5,0)
{Flickr30K Entities};

\end{axis}
\end{tikzpicture}
\vspace{18pt}

\end{minipage}%
\hfill%
% ============================================================
% RIGHT: Matching and Retrieval
% ============================================================
\begin{minipage}[t]{0.32\textwidth}
\vspace{0pt}
\centering

\textbf{(b) Matching and Retrieval}

\vspace{2pt}

\scriptsize
\renewcommand{\arraystretch}{2.3}
\setlength{\tabcolsep}{2.4pt}

\resizebox{\linewidth}{!}{%
\begin{tabular}{|ll|cccc|}
\toprule
Dataset & Model & AUC & BAcc & I@1 & T@1 \\
\midrule

VOC20
& ItemizedCLIP
& 0.752 & 0.672 & 0.389 & 0.600 \\
& \ourmodel{}
& \textbf{0.836}
& \textbf{0.746}
& \textbf{0.487}
& \textbf{0.800} \\

\midrule

ADE20K
& ItemizedCLIP
& 0.697 & 0.649 & 0.257 & 0.180 \\
& \ourmodel{}
& \textbf{0.714}
& \textbf{0.657}
& \textbf{0.367}
& \textbf{0.227} \\

\midrule

COCO
& ItemizedCLIP
& 0.889
& 0.831
& \textbf{0.124}
& \textbf{0.123} \\
& \ourmodel{}
& \textbf{0.932}
& \textbf{0.862}
& 0.106
& 0.112 \\

\midrule

Flickr30K
& ItemizedCLIP
& 0.881
& 0.812
& \textbf{0.148}
& \textbf{0.165} \\
& \ourmodel{}
& \textbf{0.925}
& \textbf{0.845}
& 0.111
& 0.152 \\

\bottomrule
\end{tabular}%
}

\end{minipage}

\vspace{-25pt}

\caption{
\textbf{Spatial evidence disentanglement and matching across OOD benchmarks.}
(a) ProtoLIP improves Pointing (P) and Energy (E); error bars show
$\pm$ standard deviation over three seeds.
(b) ProtoLIP matching and retrieval.
}

\label{fig:main_results}
\vspace{-7pt}
\end{figure*}
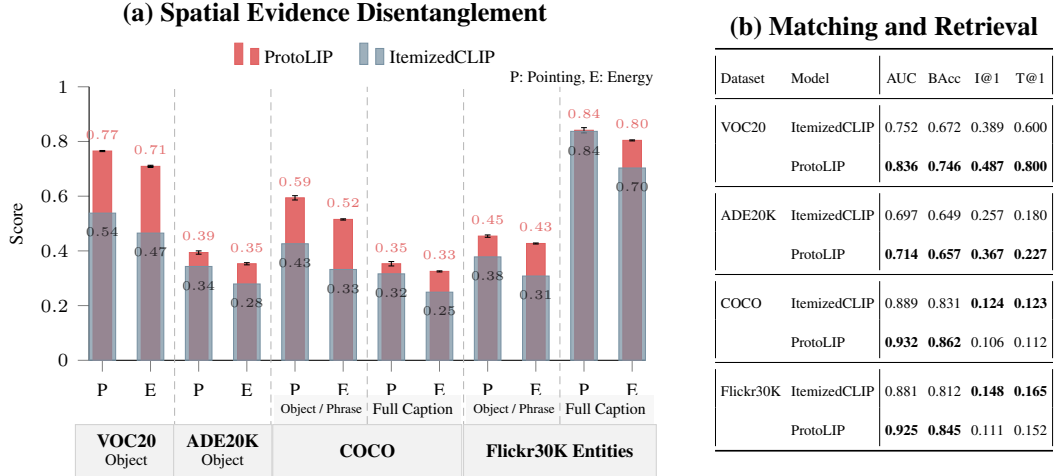

As shown in Figure~\ref{fig:main_results}, \ourmodel{} improves both Pointing and Energy on VOC20 object queries and ADE20K class queries. Gains on ADE20K are more modest, consistent with its broader long-tail vocabulary and greater semantic mismatch with the 0.3M text-only training corpus.
Nevertheless, \ourmodel{} also improves AUC and BAcc on both benchmarks,
showing that the localization gains are accompanied by stronger matching
performance.

\begin{wraptable}{r}{0.46\columnwidth}
\vspace{-10pt}
\centering
\scriptsize
\setlength{\tabcolsep}{2.2pt}
\renewcommand{\arraystretch}{1.08}
\caption{Fused ProtoLIP on Flickr30K Entities using 50/50 score-level fusion.}
\label{tab:flickr_caption_fusion}
\begin{tabular}{lccc}
\toprule
Metric pair & Backbone & ProtoLIP & Fusion \\
\midrule
Caption P/E
& 0.8367/0.7030
& \textbf{0.8437}/\textbf{0.8136}
& \textbf{0.8437}/\textbf{0.8136} \\

AUC/BAcc
& 0.8814/0.8118
& 0.9249/\textbf{0.8487}
& \textbf{0.9252}/0.8444 \\

I@1/T@1
& 0.1483/0.1653
& 0.1112/0.1523
& \textbf{0.1543}/\textbf{0.1693} \\
\bottomrule
\end{tabular}
\vspace{-10pt}
\end{wraptable}We next test whether these gains extend to richer compositional queries.
On COCO and Flickr30K, ProtoLIP further improves Pointing, Energy, AUC, and BAcc for object queries, while preserving or modestly improving
full-caption localization (Figure~\ref{fig:main_results}), which shows that stronger object-level separation does not compromise evidence
quality for complete sentence descriptions. 
Direct prototype-mediated inference improves matching separability but
moderately reduces caption retrieval. A simple 50/50 score-level fusion with
the frozen backbone not only recovers this gap but improves retrieval over the
frozen backbone, while retaining spatial evidence
(Table~\ref{tab:flickr_caption_fusion}). This separates two complementary
uses of \ourmodel{}: Direct \ourmodel{} provides a fully decomposable
prototype-mediated prediction, whereas Fused \ourmodel{} combines the
prototype-mediated and frozen-backbone scores to improve retrieval utility,
at the cost of exact decomposition of the fused score. We use Direct
\ourmodel{} as the primary model because it preserves the direct correspondence
between localized prototype evidence and the matching prediction.

\subsection{Generalization across vision–language backbones}
\label{sec:public_transfer}

We next test whether \ourmodel{} generalizes across independently pretrained
vision--language backbones. We apply the same prototype evidence layer to SigLIP-WebLI and
FLAIR merged-30M, which differ from ItemizedCLIP in both pretraining data and
objectives. For each model, the pretrained backbone remains fully frozen, and
only a new prototype evidence layer is trained on Itemized-CC0.3M. We then
evaluate the resulting models zero-shot on the same OOD benchmarks.

As shown in Table~\ref{tab:public_transfer_object_query}, \ourmodel{} improves
Pointing and Energy in all eight backbone--dataset combinations, indicating that
its localization gains are not specific to the ItemizedCLIP backbone. Matching performance also improves for
SigLIP-WebLI and remains broadly stable for FLAIR, with modest
metric-specific variation. Under the same-training-data in-domain setting,
\ourmodel{} further achieves the best AUC, BAcc, and Recall@5/10 among the
compared baselines (Appendix~\ref{app:same_data}). \begin{table*}[h]
\vspace{-15pt}
\centering
\scriptsize
\setlength{\tabcolsep}{3.0pt}
\caption{
Transfer of ProtoLIP to frozen public vision--language backbones on OOD
object- and phrase-level benchmarks. Only the ProtoLIP layer is trained
on Itemized-CC0.3M.
}
\label{tab:public_transfer_object_query}

\begin{tabular*}{\textwidth}{
@{\extracolsep{\fill}}
>{\raggedright\arraybackslash}p{1.55cm}
ll
>{\columncolor{gray!12}}c
>{\columncolor{gray!12}}c
cc
@{}
}
\toprule
Dataset & Backbone & Variant
& Pointing & Energy & AUC & BAcc \\
\midrule

COCO Object
& SigLIP-WebLI & Frozen
& 0.292 & 0.245 & 0.791 & 0.749 \\
& & + ProtoLIP Layer
& \textbf{0.390} & \textbf{0.384}
& \textbf{0.833} & \textbf{0.764} \\
\cmidrule(lr){2-7}

& FLAIR merged-30M & Frozen
& 0.245 & 0.313 & \textbf{0.906} & \textbf{0.827} \\
& & + ProtoLIP Layer
& \textbf{0.558} & \textbf{0.483}
& 0.888 & 0.804 \\

\midrule

ADE20K
& SigLIP-WebLI & Frozen
& 0.226 & 0.230 & 0.793 & 0.730 \\
& & + ProtoLIP Layer
& \textbf{0.412} & \textbf{0.363}
& \textbf{0.802} & \textbf{0.737} \\
\cmidrule(lr){2-7}

& FLAIR merged-30M & Frozen
& 0.214 & 0.278 & 0.764 & 0.691 \\
& & + ProtoLIP Layer
& \textbf{0.436} & \textbf{0.397}
& \textbf{0.785} & \textbf{0.719} \\

\midrule

VOC20
& SigLIP-WebLI & Frozen
& 0.397 & 0.365 & 0.862 & 0.828 \\
& & + ProtoLIP Layer
& \textbf{0.684} & \textbf{0.625}
& \textbf{0.912} & \textbf{0.834} \\
\cmidrule(lr){2-7}

& FLAIR merged-30M & Frozen
& 0.272 & 0.473 & \textbf{0.899} & \textbf{0.833} \\
& & + ProtoLIP Layer
& \textbf{0.802} & \textbf{0.747}
& 0.890 & 0.814 \\

\midrule

Flickr30K Entities(phrase)
& SigLIP-WebLI & Frozen
& 0.296 & 0.279 & 0.852 & 0.778 \\
& & + ProtoLIP Layer
& \textbf{0.503} & \textbf{0.451}
& \textbf{0.870} & \textbf{0.791} \\
\cmidrule(lr){2-7}

& FLAIR merged-30M & Frozen
& 0.272 & 0.354 & \textbf{0.915} & \textbf{0.837} \\
& & + ProtoLIP Layer
& \textbf{0.521} & \textbf{0.484}
& 0.898 & 0.814 \\

\bottomrule
\end{tabular*}

\vspace{-5pt}
\end{table*}

To further isolate backbone effects from differences in pretraining data, we
retrain ItemizedCLIP, Multi-positive SigLIP, FLAIR, and DreamLIP on the same
Itemized-CC0.3M data and then freeze their backbones before training ProtoLIP.
ProtoLIP improves localization for ItemizedCLIP, Multi-positive SigLIP, and
FLAIR. In particular, across the same-data retrained Multi-positive SigLIP and
FLAIR backbones, it achieves average relative gains of 66\% in Pointing and
37\% in Energy across three evaluation datasets. Localization does not improve
for DreamLIP, suggesting that transfer also depends on the underlying patch--text representations. Full matched-data results and patch--text
alignment diagnostics are provided in
Appendix~\ref{app:failure_analysis}.

\subsection{Comparison with External Localization Methods}
\vspace{-5PT}
We further ask whether the object-level evidence produced by \ourmodel{} is
competitive with existing localization approaches despite training the
prototype evidence layer without explicit spatial supervision. To test this,
we compare against complementary external baselines
(Figure~\ref{fig:public_localization_curve}): FLAIR (merged-30M) and
ItemizedCLIP as fine-grained VLM baselines, NACLIP with a CLIP backbone
pretrained on approximately 400M image--text pairs as a training-free
localization baseline, and the spatially supervised MM Grounding DINO.
All methods are re-evaluated on identical object queries and target regions.
Because MM Grounding DINO outputs scored bounding boxes rather than dense
evidence maps, we rasterize its predictions into dense spatial maps before
applying the same Pointing and Energy metrics; the rasterization protocol is
detailed in Appendix~\ref{app:implementation}.

Despite training the prototype evidence layer only on 0.3M image--text pairs, \ourmodel{} outperforms both fine-grained VLM
baselines on all eight comparisons and exceeds NACLIP on six of eight,
trailing it only on ADE20K. The weaker relative performance on ADE20K is
consistent with its broader 150-class vocabulary and greater mismatch with
the Itemized-CC0.3M training distribution. Against MM Grounding DINO,
\ourmodel{} is stronger on COCO and VOC20 under both metrics, while trailing
on ADE20K and Flickr30K. Notably, the MM Grounding DINO Large-All
checkpoint is trained with explicit spatial supervision from multiple
detection and grounding datasets, including COCO and Flickr30K. Thus,
\ourmodel{} remains competitive with a dedicated grounding model benefiting
from direct spatial supervision and training-data overlap with two of the
evaluation benchmarks.

\usetikzlibrary{patterns}

\pgfdeclarepatternformonly{dashed north east lines}
{\pgfqpoint{-1pt}{-1pt}}
{\pgfqpoint{5pt}{5pt}}
{\pgfqpoint{4pt}{4pt}}
{
    \pgfsetlinewidth{0.35pt}
    \pgfsetdash{{1.2pt}{1.1pt}}{0pt}
    \pgfpathmoveto{\pgfqpoint{-1pt}{-1pt}}
    \pgfpathlineto{\pgfqpoint{5pt}{5pt}}
    \pgfusepath{stroke}
}

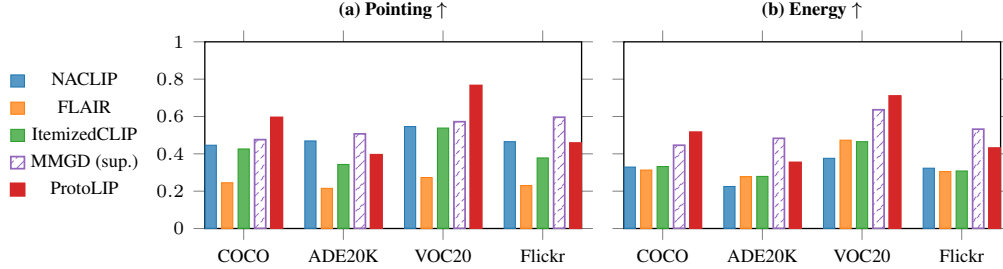
\begin{figure}[th]
\vspace{-10pt}
\centering

\definecolor{naclipcolor}{RGB}{31,119,180}
\definecolor{flaircolor}{RGB}{255,127,14}
\definecolor{itemizedcolor}{RGB}{44,160,44}
\definecolor{dinocolor}{RGB}{148,103,189}
\definecolor{ourscolor}{RGB}{214,39,40}

\begin{tikzpicture}

% ============================================================
% Pointing
% ============================================================
\begin{axis}[
name=pointing,
width=0.47\columnwidth,
height=0.29\columnwidth,
ybar,
bar width=4.2pt,
ymin=0,
ymax=1,
axis lines=box,
axis line style={line width=0.5pt},
tick align=outside,
symbolic x coords={COCO,ADE20K,VOC20,Flickr},
xtick=data,
xticklabel style={font=\scriptsize},
yticklabel style={font=\scriptsize},
ytick={0,0.2,0.4,0.6,0.8,1.0},
title={(a) Pointing $\uparrow$},
title style={font=\scriptsize\bfseries,yshift=-1pt},
enlarge x limits=0.13,
legend image code/.code={
    \draw[#1] (0cm,-0.08cm) rectangle (0.18cm,0.08cm);
},
legend style={
    at={(-0.14,0.5)},
    anchor=east,
    legend columns=1,
    font=\scriptsize,
    draw=none,
    fill=none,
    /tikz/every even column/.append style={column sep=2pt}
}
]

\addplot[fill=naclipcolor,draw=naclipcolor,fill opacity=0.75]
coordinates {
(COCO,0.446)
(ADE20K,0.469)
(VOC20,0.546)
(Flickr,0.465)
};

\addplot[fill=flaircolor,draw=flaircolor,fill opacity=0.75]
coordinates {
(COCO,0.245)
(ADE20K,0.215)
(VOC20,0.273)
(Flickr,0.230)
};

\addplot[fill=itemizedcolor,draw=itemizedcolor,fill opacity=0.75]
coordinates {
(COCO,0.426)
(ADE20K,0.343)
(VOC20,0.538)
(Flickr,0.378)
};

% Spatially supervised MM Grounding DINO: dashed diagonal fill
\addplot[
    fill=dinocolor!18,
    draw=dinocolor,
    line width=0.65pt,
    pattern=dashed north east lines,
    pattern color=dinocolor
]
coordinates {
(COCO,0.476)
(ADE20K,0.507)
(VOC20,0.572)
(Flickr,0.596)
};

\addplot[fill=ourscolor,draw=ourscolor,line width=0.8pt]
coordinates {
(COCO,0.594)
(ADE20K,0.394)
(VOC20,0.765)
(Flickr,0.457)
};

\legend{
NACLIP,
FLAIR,
ItemizedCLIP,
MMGD (sup.),
\ourmodel{}
}

\end{axis}

% ============================================================
% Energy
% ============================================================
\begin{axis}[
name=energy,
at={(pointing.east)},
anchor=west,
xshift=0.04\columnwidth,
width=0.47\columnwidth,
height=0.29\columnwidth,
ybar,
bar width=4.2pt,
ymin=0,
ymax=1,
axis lines=box,
axis line style={line width=0.5pt},
tick align=outside,
symbolic x coords={COCO,ADE20K,VOC20,Flickr},
xtick=data,
xticklabel style={font=\scriptsize},
ytick={0,0.2,0.4,0.6,0.8,1.0},
yticklabels={},
title={(b) Energy $\uparrow$},
title style={font=\scriptsize\bfseries,yshift=-1pt},
enlarge x limits=0.13
]

\addplot[fill=naclipcolor,draw=naclipcolor,fill opacity=0.75]
coordinates {
(COCO,0.329)
(ADE20K,0.225)
(VOC20,0.376)
(Flickr,0.323)
};

\addplot[fill=flaircolor,draw=flaircolor,fill opacity=0.75]
coordinates {
(COCO,0.313)
(ADE20K,0.278)
(VOC20,0.473)
(Flickr,0.305)
};

\addplot[fill=itemizedcolor,draw=itemizedcolor,fill opacity=0.75]
coordinates {
(COCO,0.332)
(ADE20K,0.279)
(VOC20,0.465)
(Flickr,0.308)
};

% Spatially supervised MM Grounding DINO: dashed diagonal fill
\addplot[
    fill=dinocolor!18,
    draw=dinocolor,
    line width=0.65pt,
    pattern=dashed north east lines,
    pattern color=dinocolor
]
coordinates {
(COCO,0.446)
(ADE20K,0.483)
(VOC20,0.636)
(Flickr,0.532)
};

\addplot[fill=ourscolor,draw=ourscolor,line width=0.8pt]
coordinates {
(COCO,0.515)
(ADE20K,0.353)
(VOC20,0.709)
(Flickr,0.430)
};

\end{axis}

\end{tikzpicture}

\vspace{-14pt}
\caption{
OOD object-level localization across fine-grained VLMs, training-free
localization, spatially supervised grounding, and \ourmodel{}.
}
\label{fig:public_localization_curve}
\vspace{-8pt}
\end{figure}

\subsection{Ablation Study}
\label{sec:ablation}

We ablate ProtoLIP on COCO to isolate the effects of prototype mediation,
semantic organization, and spatial regularization. As shown in
Table~\ref{tab:coco-mixed-query-ablation}, a flat prototype layer improves
object-level localization, while semantic family organization provides the
largest additional gain. Combining the area and overlap constraints yields
the strongest localization.

\begin{table*}[h]
\centering
\vspace{-5mm}
\scriptsize
\setlength{\tabcolsep}{3.5pt}
\caption{Ablation of ProtoLIP components on COCO object-level queries.}
\label{tab:coco-mixed-query-ablation}
\resizebox{0.8\textwidth}{!}{%
\begin{tabular}{lcccccccc}
\toprule
& & & & &
\multicolumn{2}{c}{\textbf{Object-level localization}} &
\multicolumn{2}{c}{\textbf{Matching}} \\
\cmidrule(lr){6-7}
\cmidrule(lr){8-9}
Variant & Proto. & Family & Area & Overlap
& Pointing & Energy
& AUC & BAcc \\
\midrule

Frozen ItemizedCLIP
& -- & -- & -- & --
& 0.426 & 0.332
& 0.889 & 0.831 \\

Flat Prototype
& \checkmark & -- & -- & --
& 0.492 & 0.407
& 0.919 & 0.846 \\

Family Prototype
& \checkmark & \checkmark & -- & --
& 0.586 & 0.461
& \textbf{0.934} & \underline{0.863} \\

$+$ Area
& \checkmark & \checkmark & \checkmark & --
& \underline{0.588} & \underline{0.494}
& \underline{0.933} & 0.862 \\

$+$ Overlap
& \checkmark & \checkmark & -- & \checkmark
& 0.582 & 0.459
& \textbf{0.934} & \textbf{0.864} \\

Full ProtoLIP
& \checkmark & \checkmark & \checkmark & \checkmark
& \textbf{0.594} & \textbf{0.515}
& 0.932 & 0.862 \\

\bottomrule
\end{tabular}%
}
\vspace{-10pt}
\end{table*}

ProtoLIP adds only 2.23M trainable parameters, equivalent to 1.48\% of the frozen backbone. The resulting localization gains therefore require only a small
parameter overhead. Full parameter counts are provided in
Appendix~\ref{app:param_budget}.

\vspace{-5PT}
\section{Conclusion}
\label{sec:conclusion}
\vspace{-5PT}
We show that increasingly fine-grained text conditioning does not necessarily
yield correspondingly precise object-level evidence. We introduce ProtoLIP, a
lightweight prototype-mediated evidence layer that improves query-specific
localization and separation without spatial supervision or backbone retraining.
Crucially, ProtoLIP constructs its matching score directly from localized prototype
evidence, enabling the score of a complete sentence query to be exactly
decomposed into semantic-family and prototype contributions.

\newpage
\subsection*{AI use statement}

In this work, we used generative AI tools to assist with the offline
construction of the semantic-family taxonomy, including extracting and
grouping localizable visual concepts from training descriptions. We also used
generative AI tools to assist with manuscript editing and improving
readability. All AI-assisted outputs were reviewed and verified by the
authors. We take responsibility for the final content of this work, including
text, claims, and artifacts produced with the aid of generative AI.
\bibliography{iclr2027_conference}
\bibliographystyle{iclr2027_conference}
\newpage
\appendix

% ============================================================
% LLM Prompt Templates for Taxonomy Construction
% ============================================================

\section{LLM Prompt Templates for Taxonomy Construction}
\label{app:llm_prompts}

We used the following offline prompt templates to construct the semantic
family taxonomy. The LLM was given only sampled Itemized-CC0.3M training
descriptions; no images, evaluation labels, downstream test data, or model
outputs were provided.

\par\medskip
\noindent\textbf{Prompt 1: Visual-concept extraction.}\par
\begin{lstlisting}[style=taxonomyprompt]
You are helping construct a visual concept vocabulary from image-text training descriptions.

Input: a batch of image descriptions.

Your task is to extract only explicitly mentioned visual entities or image regions that can be spatially localized in an image.

Rules:
1. Keep concrete objects, object parts, people/animals, scene regions, and visually identifiable materials or places when they are localizable.
2. Exclude actions, relations, colors, styles, emotions, subjective attributes, quantities alone, and abstract concepts.
3. Normalize obvious lexical variants to one canonical concept.
4. Record aliases only if they appear in the provided descriptions.
5. Do not introduce concepts or aliases unsupported by the input.
6. If a phrase is too vague or not spatially localizable, discard it.

Return JSON:
{
  "concepts": [
    {
      "canonical": "canonical concept name",
      "aliases": ["alias 1", "alias 2"],
      "evidence_descriptions": [0, 3]
    }
  ]
}

The evidence_descriptions field lists the indices of input descriptions where the concept was observed.

Input descriptions:
[BATCH_OF_DESCRIPTIONS]
\end{lstlisting}

\par\medskip
\noindent\textbf{Prompt 2: Adaptive semantic grouping.}\par
\begin{lstlisting}[style=taxonomyprompt]
You are helping organize a visual concept vocabulary into coarse semantic families for routing visual prototypes.

Input: a list of canonical visual concepts and their observed aliases.

Your task is to group concepts into semantic families that are visually and semantically coherent.

Rules:
1. Choose the number of families based on the concept inventory.
2. Families should be coarse routing groups, not fine-grained class labels.
3. Separate concepts with substantially different visual structure or role.
4. Merge lexical variants and close synonyms into the same family.
5. Allow multiple-family membership only for genuinely ambiguous concepts.
6. Do not create families for actions, relations, colors, or styles.
7. Do not add concepts or aliases absent from the input inventory.

Return JSON:
{
  "families": [
    {
      "family": "family name",
      "description": "short visual definition",
      "concepts": [
        "canonical concept 1",
        "canonical concept 2"
      ],
      "aliases": ["alias 1", "alias 2"]
    }
  ]
}

Concept inventory:
[CONCEPT_INVENTORY]
\end{lstlisting}

\par\medskip
\noindent\textbf{Prompt 3: Taxonomy audit.}\par
\begin{lstlisting}[style=taxonomyprompt]
You are auditing a draft semantic taxonomy for visual prototype routing.

Input: a draft family-to-alias taxonomy and the original concept inventory.

Your task is to check whether the taxonomy is complete, consistent, and supported by the training-text inventory.

Audit requirements:
1. Identify missing concepts that should be assigned to a family.
2. Identify duplicate aliases or inconsistent family assignments.
3. Identify visually incoherent families.
4. Identify aliases or concepts unsupported by the input inventory.
5. Identify overly broad or overly narrow families.
6. Preserve coarse semantic routing.
7. Base every finding and revision only on the provided concept inventory; do not introduce unsupported concepts.

Return JSON:
{
  "issues": [
    {
      "type": "missing | duplicate | ambiguous | incoherent | unsupported",
      "item": "concept or alias",
      "explanation": "brief reason",
      "suggested_fix": "specific correction"
    }
  ],
  "revised_taxonomy": [
    {
      "family": "family name",
      "concepts": [
        "canonical concept 1",
        "canonical concept 2"
      ],
      "aliases": ["alias 1", "alias 2"]
    }
  ]
}

Draft taxonomy:
[DRAFT_TAXONOMY]

Concept inventory:
[CONCEPT_INVENTORY]
\end{lstlisting}

\section{Evaluation Protocol and Implementation Details}
\label{app:implementation}

\paragraph{Prototype layer architecture.}
In our implementation, $\phi_\theta$ is a gated two-layer residual MLP
with hidden width $d/2$, where $d$ is the frozen backbone's patch-embedding
dimension. We use GELU activation, a residual scale of $\gamma=0.25$, and
final $\ell_2$ normalization.

\paragraph{Prototype initialization.}
Before training, the $M=512$ prototype vectors are independently sampled
from a standard Gaussian distribution and $\ell_2$-normalized,
\[
p_k \sim \mathcal{N}(0,I),
\qquad
p_k \leftarrow \frac{p_k}{\|p_k\|_2}.
\]
We then project each prototype onto the training patch-feature space.
Using patch features from up to 8,192 randomly sampled training image--text
feature rows, each prototype is replaced by the $\ell_2$-normalized mean of
its 128 most similar patch features. The resulting prototypes are subsequently
optimized as learnable parameters throughout training. This initialization is
independent of the semantic-family construction: the K23/K39 taxonomy
determines only prototype-bank membership and query routing. No downstream
evaluation data, spatial annotations, K-means clustering, or text embeddings
are used for prototype initialization.

\paragraph{Optimization and hyperparameters.}
We optimize the prototype evidence layer with AdamW for 6,000 steps using a
learning rate of $1.5\times10^{-3}$ and weight decay of $10^{-3}$. The
effective batch size is 1,024. Distributed runs use two NVIDIA GPUs with 256
samples per GPU and two gradient-accumulation steps; single-GPU sensitivity
experiments use a batch size of 256 with four accumulation steps. Unless
otherwise specified, reported \ourmodel{} results are averaged over three
independent training runs with random seeds 1713, 1714, and 1715.

The prototype and matching temperatures are set to $0.04$ and $0.05$,
respectively. Image patches are pooled with a $2\times2$ region size. We set
$\lambda_{\mathrm{area}}=1.5$, $\tau_a=0.14$,
$\lambda_{\mathrm{overlap}}=0.25$, and $\tau_o=0.20$.
Gradient checkpointing and chunked score computation are used to reduce
memory consumption without changing the model or objective. The frozen
vision--language backbone is shared across all variants.

\paragraph{External baseline checkpoints.}

We evaluate the public
(openmmlab-community/mm\_grounding\_dino\_large\_all) checkpoint.
Its training corpus includes GoldG, V3Det, COCO2017, LVIS-v1, COCO2014,
GRIT, RefCOCO, RefCOCO+, RefCOCOg, and gRefCOCO. NACLIP uses the public
CLIP-B/16 checkpoint, while FLAIR uses its released merged-30M checkpoint.

For MM Grounding DINO, each query is evaluated independently using the
single-query prompt followed by a period. We use box and text confidence
thresholds of $0.25$ and retain the bounding boxes and corresponding detection
scores returned by the model's post-processing procedure. To obtain a dense
map compatible with Pointing and Energy, each retained detection score is
added to every cell of a $64\times64$ grid whose center lies inside the
corresponding predicted box. Scores from overlapping boxes are summed, and
the resulting heatmap is normalized by its total mass. Pointing is computed
from the maximum-mass cell, while Energy is the normalized heatmap mass
inside the ground-truth region.

This conversion preserves the detector's predicted regions and returned
scores while providing a common spatial representation for evaluation.
MM Grounding DINO is spatially supervised with detection and
referring-expression annotations, whereas ProtoLIP receives no spatial
annotations and keeps its VLM backbone frozen. We therefore treat this
comparison as a reference against supervised grounding.

\paragraph{COCO caption-linked object queries.}
COCO does not provide phrase--box correspondences, so we construct a fixed
caption-linked object-query manifest from validation captions. Before model
evaluation, we define a boundary-aware alias list for COCO instance categories,
e.g., \textit{bike} for \textit{bicycle} and \textit{woman}/\textit{man} for
\textit{person}, and apply deterministic lexical matching to each caption. We
retain only matched categories with at least one ground-truth instance box in
the image, discarding unmatched, non-localizable, or unannotated mentions. If
multiple instances of a matched category are present, their boxes form a union
target: Pointing is correct when the top evidence patch falls inside any
instance box, and Energy is the evidence mass within their union. The resulting
manifest contains 681 object queries over 557 images and is shared by all
methods.

\subsection{Additional COCO-80 Evaluation}
\label{app:coco80}

We additionally evaluate \ourmodel{} on the complete set of 80 COCO object
categories. This evaluation complements the caption-linked COCO object-query
benchmark in the main paper by covering the full COCO-80 category set.
As shown in Table~\ref{tab:coco80}, \ourmodel{} improves both Pointing and
Energy over the frozen ItemizedCLIP backbone.

\begin{table}[h]
\centering
\small
\setlength{\tabcolsep}{6pt}
\caption{
Object-level evidence localization on the complete COCO-80 category set.
}
\label{tab:coco80}
\begin{tabular}{lcc}
\toprule
Model & Pointing & Energy \\
\midrule
ItemizedCLIP & 0.313 & 0.250 \\
\ourmodel{}  & \textbf{0.389} & \textbf{0.358} \\
\midrule
Improvement  & +0.075 & +0.108 \\
\bottomrule
\end{tabular}
\end{table}

\section{Natural Fallback and Semantic-Family Coverage}
\label{app:natural_fallback}

Our semantic-family taxonomy provides a coarse routing prior rather than a
closed vocabulary. When a query matches one or more semantic families, only
prototypes from the corresponding family banks are eligible for scoring.
When no family is matched, ProtoLIP falls back to the full prototype pool,
allowing arbitrary queries to remain supported without requiring an explicit
taxonomy assignment.

We further examine the coverage of semantic-family routing and the behavior
of this fallback on two broader-vocabulary benchmarks: ADE20K, which contains
a challenging long-tail object vocabulary, and Flickr30K Entities, which
contains open-vocabulary phrase-level queries. We report both unique-query
coverage, which measures the fraction of distinct queries matched by the
taxonomy, and sample coverage, which measures the fraction of evaluation
samples for which semantic-family routing is active. We additionally split
the localization results according to whether each query is matched or
unmatched and report the improvement of ProtoLIP over the frozen ItemizedCLIP
backbone.

\begin{table}[t]
\centering
\caption{
Semantic-family routing coverage and matched/unmatched localization analysis.
$\Delta$P and $\Delta$E denote the absolute improvements of ProtoLIP over the
frozen ItemizedCLIP backbone in Pointing and Energy, respectively.
}
\label{tab:routing_coverage}
\small
\begin{tabular}{lcccc}
\toprule
Dataset
& Unique-query cov.
& Sample cov.
& Matched $\Delta$P / $\Delta$E
& Unmatched $\Delta$P / $\Delta$E \\
\midrule
ADE20K
& 50.00\%
& 72.59\%
& +0.071 / +0.093
& $-$0.007 / +0.016 \\
Flickr30K Entities
& 63.76\%
& 73.26\%
& +0.103 / +0.154
& +0.013 / +0.033 \\
\bottomrule
\end{tabular}
\end{table}

As shown in Table~\ref{tab:routing_coverage}, the largest localization gains
occur on matched queries, where semantic-family routing constrains prototype
eligibility. For unmatched queries, the all-prototype fallback largely
preserves localization performance: Pointing changes are close to zero,
while Energy shows modest improvements on both benchmarks. These results
indicate that semantic-family routing provides the primary localization
benefit when applicable, while the fallback maintains support for queries
outside the taxonomy without substantial degradation.

\subsection{Behavior on Naturally Unmatched Queries}
\label{app:natural_fallback_indomain}

We next examine whether queries that fail to match any semantic family
are systematically degraded. This analysis uses the
held-out in-domain Itemized-CC0.3M split and is distinct from the OOD
benchmarks evaluated in the main paper.

Reported values are mean $\pm$ sample standard deviation over three
independent training runs with different random seeds. The same evaluation
protocol is applied to each trained model. Results are reported in
Table~\ref{tab:natural_fallback}.

\begin{table}[h]
\centering
\scriptsize
\caption{
Performance on naturally matched and unmatched Itemized-CC0.3M queries.
Values are mean $\pm$ sample standard deviation over three seeds. Unmatched
queries use the complete prototype-bank fallback.
}
\label{tab:natural_fallback}
\begin{tabular}{lcccc}
\toprule
Query subset / routing & AUC & BAcc & I@1 & T@1 \\
\midrule

Matched, standard routing
& 0.999$\pm$0.000
& 0.989$\pm$0.001
& 0.746$\pm$0.010
& 0.798$\pm$0.012 \\

Matched, forced fallback
& 0.999$\pm$0.000
& 0.987$\pm$0.001
& 0.795$\pm$0.018
& 0.815$\pm$0.009 \\

Naturally unmatched, fallback
& 0.998$\pm$0.000
& 0.986$\pm$0.002
& 0.782$\pm$0.010
& 0.777$\pm$0.013 \\

\bottomrule
\end{tabular}
\end{table}

As shown in Table~\ref{tab:natural_fallback}, naturally unmatched queries
retain AUC and balanced accuracy close to those of matched queries. For these
queries, standard inference and explicitly forced fallback are identical
because unmatched queries already use the complete prototype bank.

As a controlled intervention, we additionally force matched queries to use
the complete bank instead of their routed families. Balanced accuracy remains
essentially unchanged, while I@1 increases. Matched and naturally unmatched
queries are semantically different populations by construction, so direct
comparisons between these subsets should be interpreted descriptively rather
than causally.

This same direction is visible under a separate, model-level comparison:
relative to the frozen ItemizedCLIP baseline under the identical in-domain
FullPlus protocol (Table~\ref{tab:indomain_same_data}), ProtoLIP's strict I@1
is also lower despite leading on AUC, BAcc, and broader retrieval cutoffs.
Together, these two independent in-domain comparisons suggest that
family-constrained routing carries a consistent, modest cost to strict top-1
retrieval, traded for the localization, separation, and matching gains
reported throughout the paper.

Overall, the fallback mechanism prevents unmatched queries from being
silently discarded or assigned to arbitrary semantic families.

\section{Taxonomy Granularity Analysis}
\label{app:taxonomy_granularity}

We compare the main 39-family taxonomy (K39), constructed from the
expanded extraction, with a smaller 23-family taxonomy (K23) built from
30K sampled training descriptions. Only the \ourmodel{} layer is retrained,
and both variants are evaluated under the same held-out Itemized-CC0.3M
split used throughout the paper, reported
here for the full 1,200-pair evaluation set. Results are shown
in Table~\ref{tab:taxonomy_indomain}.

\begin{table}[h]
\centering
\small
\setlength{\tabcolsep}{4pt}
\caption{
In-domain comparison between the 23-family taxonomy
(K23) and the finer 39-family taxonomy (K39). Both models use direct
prototype-mediated inference without knowledge distillation and are
evaluated on the same 1,200-pair protocol.
}
\label{tab:taxonomy_indomain}
\begin{tabular}{lcccccccc}
\toprule
Model
& AUC & BAcc
& I@1 & I@5 & I@10
& T@1 & T@5 & T@10 \\
\midrule

\ourmodel{} K23
& 0.956 & \textbf{0.881}
& 0.340 & 0.466 & 0.546
& 0.323 & 0.448 & 0.522 \\

\ourmodel{} K39
& \textbf{0.958} & 0.880
& \textbf{0.347} & \textbf{0.485} & \textbf{0.553}
& \textbf{0.337} & \textbf{0.463} & \textbf{0.526} \\

\bottomrule
\end{tabular}
\end{table}

\begin{table}[h]
\centering
\scriptsize
\caption{
Effect of taxonomy granularity. K23 and K39 denote the 23-family and
39-family taxonomies, respectively; both use direct prototype-mediated
inference without knowledge distillation. Differences between the
two taxonomies are within one point on all reported metrics.
}
\label{tab:taxonomy_granularity}
\begin{tabular}{lcccc}
\toprule
& \multicolumn{2}{c}{K23} & \multicolumn{2}{c}{K39} \\
\cmidrule(lr){2-3}\cmidrule(lr){4-5}
Dataset / Query & Pointing & Energy & Pointing & Energy \\
\midrule
COCO object / phrase & .5868 & .5187 & .5939 & .5143 \\
Flickr30K phrase & .4553 & .4293 & .4571 & .4301 \\
\bottomrule
\end{tabular}
\end{table}

Table~\ref{tab:taxonomy_indomain} shows that increasing the number of
semantic families from 23 to 39 produces only modest changes in matching
and retrieval. K39 slightly improves AUC and several retrieval cutoffs,
while BAcc remains essentially unchanged. Table~\ref{tab:taxonomy_granularity}
shows that K39 additionally yields modestly stronger spatial localization,
with differences from K23 within one point on all reported metrics. We
adopt K39 as our main taxonomy for this modest localization benefit and
its broader semantic coverage; the small gap to K23 on both matching and
localization indicates that our results are not sensitive to this
granularity choice.

\section{Parameter Budget}
\label{app:param_budget}

To verify that the improvements in the main ablation are not caused by
differences in trainable capacity, we report the parameter budget of all
prototype variants in Table~\ref{tab:param_budget}.

\begin{table}[h]
\centering
\small
\caption{
Parameter budget across ablation variants. The ItemizedCLIP backbone remains
frozen, and only the ProtoLIP layer is optimized. All prototype variants
introduce the same 2.23M trainable parameters (1.48\% of the backbone).
Family routing uses fixed masks, while the area and overlap objectives are
parameter-free.
}
\label{tab:param_budget}
\begin{tabular}{lrrrr}
\toprule
Model & Frozen Backbone & Trainable & Total & Increase \\
\midrule

Frozen ItemizedCLIP
& 150.686M & 0 & 150.686M & 0.00\% \\

Flat Prototype
& 150.686M & 2.230M & 152.916M & 1.48\% \\

Family Prototype
& 150.686M & 2.230M & 152.916M & 1.48\% \\

Full ProtoLIP
& 150.686M & 2.230M & 152.916M & 1.48\% \\

\bottomrule
\end{tabular}
\end{table}

Table~\ref{tab:param_budget} confirms that Flat Prototype, Family Prototype,
and Full ProtoLIP use identical trainable parameter budgets. Improvements
from semantic family routing and spatial regularization therefore cannot be
attributed to increased model capacity.

\section{Same-Data In-Domain Controls}
\label{app:same_data}

We additionally compare ProtoLIP with alternative vision--language objectives
trained using the same Itemized-CC0.3M training manifest. This controls for
differences in pretraining data that affect comparisons with released
large-scale checkpoints. Results are reported in
Table~\ref{tab:indomain_same_data}.

\begin{table*}[h]
\centering
\small
\setlength{\tabcolsep}{4pt}
\caption{
In-domain image--text matching and retrieval on the same-data FullPlus test
set with 1,200 image--text pairs. All transferred models are trained on
Itemized-CC0.3M. \ourmodel{} uses the direct prototype-mediated score without
knowledge distillation or inference-time backbone fusion. I@K and T@K denote
image-to-text and text-to-image Recall@K, respectively. Best results are shown
in \textbf{bold}; second-best results are \underline{underlined}.
}
\label{tab:indomain_same_data}
\resizebox{\textwidth}{!}{%
\begin{tabular}{lcccccccc}
\toprule
Model
& AUC & BAcc
& I@1 & I@5 & I@10
& T@1 & T@5 & T@10 \\
\midrule

ItemizedCLIP
& 0.845 & 0.779
& \textbf{0.368} & 0.406 & 0.434
& \textbf{0.356} & \underline{0.408} & 0.438 \\

Single-positive SigLIP-0.3M
& \underline{0.908} & \underline{0.820}
& 0.263 & 0.375 & 0.428
& 0.223 & 0.355 & 0.398 \\

Multi-positive SigLIP-0.3M
& 0.905 & \underline{0.820}
& 0.334 & \underline{0.422} & \underline{0.466}
& 0.306 & 0.403 & 0.437 \\

FLAIR-0.3M
& 0.892 & 0.804
& 0.323 & 0.417 & 0.455
& 0.290 & 0.399 & \underline{0.441} \\

DreamLIP-0.3M
& 0.829 & 0.763
& 0.100 & 0.203 & 0.264
& 0.073 & 0.168 & 0.253 \\

\ourmodel{}
& \textbf{0.958} & \textbf{0.880}
& \underline{0.347} & \textbf{0.485} & \textbf{0.553}
& \underline{0.337} & \textbf{0.463} & \textbf{0.526} \\

\bottomrule
\end{tabular}%
}
\end{table*}

As shown in Table~\ref{tab:indomain_same_data}, ProtoLIP achieves the strongest
AUC and balanced accuracy among the same-data controls. Although ItemizedCLIP
retains the highest strict Recall@1, ProtoLIP provides the strongest retrieval
performance at several broader cutoffs, including I@5, I@10, T@5, and T@10.
These results indicate that the gains in matching quality are not explained
solely by differences in pretraining data or scale.

\section{Matched-Data Backbone Transfer and Patch--Text Alignment}
\label{app:failure_analysis}

To separate backbone effects from pretraining-data differences, we evaluate
\ourmodel{} on backbones trained on the same Itemized-CC0.3M corpus. The
results in Table~\ref{tab:matched_backbone_transfer_full} show that transfer
remains backbone-dependent. We hypothesize that \ourmodel{} requires frozen
patch tokens that are meaningfully aligned with the text space, since its
matching score is constructed entirely from prototype-pooled patch evidence.

\begin{table*}[t]
\centering
\scriptsize
\setlength{\tabcolsep}{4.5pt}
\caption{
Matched-data transfer across frozen vision--language backbones under the full
evaluation protocol. All backbones are trained on Itemized-CC0.3M and then
frozen; only the ProtoLIP layer is trained. COCO uses the object-query
benchmark, while ADE20K and VOC20 use full validation splits.
ItemizedCLIP-0.3M rows report three-seed means (seeds 1713--1715); the
remaining backbones report a single run.
}
\label{tab:matched_backbone_transfer_full}
\resizebox{\textwidth}{!}{%
\begin{tabular}{lllcccc}
\toprule
Dataset & Frozen backbone & Variant
& Pointing & Energy & AUC & BAcc \\
\midrule

COCO object
& ItemizedCLIP-0.3M & Frozen
& .426 & .332 & .889 & .831 \\
& & + ProtoLIP
& \textbf{.594} & \textbf{.515} & \textbf{.932} & \textbf{.862} \\
\cmidrule(lr){2-7}

& Multi-positive SigLIP-0.3M & Frozen
& .185 & .210 & \textbf{.821} & \textbf{.750} \\
& & + ProtoLIP
& \textbf{.456} & \textbf{.374} & .711 & .659 \\
\cmidrule(lr){2-7}

& FLAIR-0.3M & Frozen
& .243 & .260 & \textbf{.793} & \textbf{.729} \\
& & + ProtoLIP
& \textbf{.329} & \textbf{.315} & \textbf{.793} & .721 \\
\cmidrule(lr){2-7}

& DreamLIP-0.3M & Frozen
& \textbf{.306} & \textbf{.298} & \textbf{.791} & \textbf{.729} \\
& & + ProtoLIP
& .262 & .283 & .731 & .676 \\

\midrule

VOC20 full val
& ItemizedCLIP-0.3M & Frozen
& .538 & .465 & .752 & .672 \\
& & + ProtoLIP
& \textbf{.765} & \textbf{.709} & \textbf{.836} & \textbf{.746} \\
\cmidrule(lr){2-7}

& Multi-positive SigLIP-0.3M & Frozen
& .288 & .314 & \textbf{.872} & \textbf{.798} \\
& & + ProtoLIP
& \textbf{.654} & \textbf{.543} & .801 & .734 \\
\cmidrule(lr){2-7}

& FLAIR-0.3M & Frozen
& .399 & .408 & \textbf{.850} & \textbf{.778} \\
& & + ProtoLIP
& \textbf{.522} & \textbf{.487} & .845 & .758 \\
\cmidrule(lr){2-7}

& DreamLIP-0.3M & Frozen
& \textbf{.515} & \textbf{.440} & \textbf{.811} & \textbf{.739} \\
& & + ProtoLIP
& .438 & .424 & .765 & .702 \\

\midrule

ADE20K full val
& ItemizedCLIP-0.3M & Frozen
& .342 & .279 & .697 & .649 \\
& & + ProtoLIP
& \textbf{.394} & \textbf{.353} & \textbf{.714} & \textbf{.657} \\
\cmidrule(lr){2-7}

& Multi-positive SigLIP-0.3M & Frozen
& .212 & .225 & .685 & .634 \\
& & + ProtoLIP
& \textbf{.303} & \textbf{.275} & \textbf{.685} & \textbf{.637} \\
\cmidrule(lr){2-7}

& FLAIR-0.3M & Frozen
& .248 & .252 & .644 & .602 \\
& & + ProtoLIP
& \textbf{.283} & \textbf{.271} & \textbf{.695} & \textbf{.642} \\
\cmidrule(lr){2-7}

& DreamLIP-0.3M & Frozen
& \textbf{.257} & \textbf{.244} & .721 & .673 \\
& & + ProtoLIP
& .253 & .237 & \textbf{.730} & \textbf{.676} \\

\bottomrule
\end{tabular}%
}
\end{table*}

We test this property on a shared set of 400 held-out image--text pairs.
Patch Top-8 scores are computed by averaging the eight highest patch--text
cosine similarities, without prototype pooling or family routing. Patch
Top-8 AUC measures pairwise discrimination, while Patch I@1 measures
image-to-text retrieval over the 400 candidates.

\begin{table}[h]
\centering
\small
\caption{
Patch--text alignment on the shared 400-pair diagnostic set.
These diagnostic values are separate from the formal benchmark
results in Table~\ref{tab:matched_backbone_transfer_full}.
}
\label{tab:patch_alignment_diagnostics}
\begin{tabular}{lccc}
\toprule
Backbone & Patch Top-8 AUC & Patch I@1 & Routed ProtoLIP AUC \\
\midrule
DreamLIP-0.3M              & 0.513 & 0.003 & 0.845 \\
Multi-positive SigLIP-0.3M & 0.496 & 0.000 & 0.880 \\
FLAIR-0.3M                 & 0.782 & 0.308 & 0.953 \\
\bottomrule
\end{tabular}
\end{table}

Patch Top-8 AUC near $0.5$ and Patch I@1 near the $1/400 = 0.0025$ chance
level indicate that the patch tokens of DreamLIP and Multi-positive SigLIP
carry almost no text-discriminative information on their own, whereas FLAIR
exposes substantially more ($0.782$ AUC, $0.308$ Patch I@1). These backbones
remain strong at the image level
(Table~\ref{tab:matched_backbone_transfer_full}), suggesting that
the missing discriminability is concentrated in their local patch
representations.

Since the ProtoLIP score is constructed entirely from prototype-pooled patch
evidence, matching performance is retained only where such alignment is
present, consistent with the more stable score-level transfer observed on
FLAIR. Disabling family routing does not recover the lost matching performance,
suggesting that weak patch--text alignment, rather than taxonomy mismatch, is
the limiting factor.

\begin{table}[t]
\centering
\small
\setlength{\tabcolsep}{4.5pt}
\caption{
Average change from adding a ProtoLIP layer to matched-data frozen backbones
across COCO object, ADE20K full val, and VOC20 full val. Localization gains
and matching retention behave differently. Computed from
Table~\ref{tab:matched_backbone_transfer_full}.
}
\label{tab:backbone_delta_diagnosis}
\begin{tabular}{lcccc}
\toprule
Backbone & $\Delta$Pointing & $\Delta$Energy & $\Delta$AUC & $\Delta$I@1 \\
\midrule
ItemizedCLIP-0.3M
& $+0.149$ & $+0.167$ & $+0.048$ & $+0.059$ \\
Multi-positive SigLIP-0.3M
& $\mathbf{+0.243}$ & $+0.148$ & $-0.060$ & $-0.177$ \\
FLAIR-0.3M
& $+0.081$ & $+0.051$ & $+0.015$ & $-0.023$ \\
DreamLIP-0.3M
& $-0.042$ & $-0.013$ & $-0.032$ & $-0.129$ \\
\bottomrule
\end{tabular}
\end{table}

Table~\ref{tab:backbone_delta_diagnosis} summarizes the average change from
adding a ProtoLIP layer. Localization and score-level matching do not move
together: Multi-positive SigLIP-0.3M gains the most in Pointing ($+0.243$)
while losing the most in retrieval ($-0.177$), whereas DreamLIP-0.3M degrades
on both. Patch--text alignment
(Table~\ref{tab:patch_alignment_diagnostics}) tracks the matching columns
across all four backbones, but does not separate DreamLIP-0.3M from
Multi-positive SigLIP-0.3M, whose alignment is comparable ($0.513$ vs
$0.496$) while their localization moves in opposite directions. We therefore
report localization transfer and score-level matching transfer as separate
empirical behaviors, and leave the determinants of the former to future work.

\end{document}